\documentclass[10pt,letterpaper]{article}
\usepackage[top=0.85in,left=2.75in,footskip=0.75in]{geometry}

\usepackage{amsmath,amssymb}

\usepackage{changepage}

\usepackage{textcomp,marvosym}

\usepackage{cite}

\usepackage{nameref,hyperref}

\usepackage[right]{lineno}

\usepackage[nopatch=eqnum]{microtype}
\DisableLigatures[f]{encoding = *, family = * }

\usepackage[table]{xcolor}

\usepackage{array}

\usepackage{booktabs}

\newcolumntype{+}{!{\vrule width 2pt}}

\newlength\savedwidth

\raggedright
\usepackage[aboveskip=1pt,labelfont=bf,labelsep=period,justification=raggedright,singlelinecheck=off]{caption}

\usepackage[normalem]{ulem}
\usepackage{xcolor}

\makeatletter
\renewcommand{\@biblabel}[1]{\quad#1.}
\makeatother

\usepackage{lastpage,fancyhdr,graphicx}
\usepackage{epstopdf}
\fancyheadoffset[L]{2.25in}
\fancyfootoffset[L]{2.25in}
\begin{document}
\vspace*{0.2in}

\begin{flushleft}
{\Large
\textbf{Automated identification of ivory carvings using machine learning: A proof-of-concept approach to detecting the illegal wildlife trade}
}
\newline
\\
Andrius Tamosiunas\textsuperscript{3,4},
Simone Mozzon\textsuperscript{1},
David Bacon\textsuperscript{1},
Caroline Cox\textsuperscript{2},
Alan Collins\textsuperscript{5}
\\
\bigskip
\textbf{1} Institute of Cosmology and Gravitation, University of Portsmouth, Dennis Sciama Building, Burnaby Road, Portsmouth, PO1 3FX, UK
\\
\textbf{2} School of Law, Faculty of Business and Law, University of Portsmouth, Richmond Building, Portland St, Portsmouth, PO1 3DE, UK
\\
\textbf{3} Institute of Theoretical Astrophysics, University of Oslo, PO Box 1029, Blindern 0315, Norway
\\
\textbf{4} CERCA/ISO, Department of Physics, Case Western Reserve University, 10900 Euclid Avenue, Cleveland, OH, 44106, USA
\\
\textbf{5} Nottingham Business School, Nottingham Trent University, 50 Shakespeare St, Nottingham, NG1 4FQ, UK
\\
\bigskip

* \href{mailto:andrius.tamosiunas@astro.uio.no}{andrius.tamosiunas@astro.uio.no}

\end{flushleft}


\section*{Abstract}
Despite a ban on the international trading of recently poached ivory, elephants are still killed for their tusks. Ivory and its products are mainly sold in online marketplaces. However, due to the huge volume and variety of adverts and the similarity of ivory with other materials that can be legally sold, it is hard for enforcement officials to identify illicit trading and prosecute sellers. Existing techniques for identifying elephant ivory, such as Fourier-transform infrared spectroscopy and microscopy, can reliably distinguish ivory from other materials, but require physical access to the item and cannot be applied to photographs alone. As a result, these methods are unsuitable for screening the large volume of images posted in online marketplaces, leaving a gap for automated, image-based identification tools. In this proof of concept work, we investigate the potential of machine learning techniques for image analysis to automatically distinguish ivory objects from bone. We employ a semantic segmentation framework using a U-Net architecture with a ResNet34 backbone, leveraging transfer learning on an augmented dataset of 541 original images. Using a dataset of images of ivory and items resembling carved ivory, several models were trialled. On a distinct test set, the classifier achieved an Area Under the Curve (AUC) of 0.81 and an overall classification accuracy of 81\% on confident samples. The paper presents the best-performing model as a proof of concept for developing a system that can effectively identify ivory and its products when indicative features of elephant ivory are present in an image. This work provides initial steps towards a machine learning-based framework designed to advise law-enforcement in detecting illegal elephant ivory trade.

\clearpage
\newgeometry{top=0.85in,left=1in,right=1in,footskip=0.75in}

\section*{Introduction}

For hundreds of years, African and Asian elephants have been hunted for their ivory, exposing their populations to the risk of extinction. The Greeks and Romans were responsible for the extinction of the North African elephant more than two millennia ago with their use of the animals for warfare, ornament and as exotic entertainment to be slain by gladiators in the Roman arenas \cite{alchin}. The Chinese elephant had met a similar fate by 600 B.C. for similar reasons \cite{LI201284}. Given humans' history of hunting elephants, it is not surprising that conservationists fear for the fate of current wild elephant populations, with recently reported figures from the World Wildlife Fund stating that wild elephant numbers have declined from 1.3 million in 1979 to 600,000 in 1989 to an estimated 400,000 today \cite{WWF}. In 1975, the risk of losing one of the world's most iconic animals saw the Convention on International Trade in Endangered Species of Wild Fauna and Flora (``CITES") add both the African Elephant (\textit{Loxodonta africana}) and the Asian Elephant (\textit{Elephas maximus}) to its Appendix I list --- indicating the species was at risk of extinction \cite{cites2021}. The Convention on International Trade in Endangered Species of Wild Fauna and Flora is an international agreement between nation states that aims to ensure that international trade in specimens of wild animals and plants does not threaten the survival of a species. Currently, there are 183 signatories, including the United Kingdom.

CITES works through its establishment of an international regulatory framework for the prevention or regulation (depending on the specific species) of trade in endangered species \cite{Wijnstekers}. Although CITES is legally binding on the Parties to the Convention, it has to be implemented by them at their own national level. The Convention works by providing a framework for the Parties and requires them to adopt their own domestic legislation to ensure that CITES is implemented at the national level of each of the member Parties \cite{cites2022}.

The Ivory Act 2018 came into force in the United Kingdom on the 6th June 2022 making the trade in ivory almost entirely illegal. With the exception of five categories of exemptions (pre-1947 ivory items of low ivory content, pre-1975 musical instruments containing ivory, pre-1918 portrait miniatures, ivory acquisitions by qualifying museums and pre-1918 ivory items of outstanding artistic importance) it is no longer legal to buy or sell ivory in the United Kingdom. Nonetheless, trading in ivory products has been found to continue on online marketplaces, such as eBay \cite{doi:10.1177/1940082920974604}.

In 2008, eBay instituted a global ban prohibiting the sale of all types of ivory (elephant, hippopotamus, walrus, narwhal) \cite{NYT2008}. However, the ban excluded the sale of bones from non-ivory-producing animals, provided the species is clearly stated in the listing description. As a result, sellers wishing to advertise ivory for sale can still breach the rules by simply listing the item with misleading descriptions, such as ``cow bone" and ``bovine bone" \cite{Cox_2020}. Most of these adverts include a number of high resolution pictures of the item which allow buyers to identify the indicative features of elephant ivory, like the presence of Schreger lines\footnote{Schreger lines are distinctive, cross-hatching structural patterns visible in the cross-sections of elephant and mammoth ivory, used by analysts to identify genuine ivory and differentiate between species based on the angle at which the lines intersect.} and the quality of the carving \cite{Cox_2020}.

Despite these stringent domestic frameworks, the global illegal ivory trade remains resilient, driven by high demand in persistent illicit hubs. To enforce these bans, forensic specialists rely on established, definitive laboratory methods to identify genuine ivory and its products. These primarily include Fourier-transform infrared (FTIR) spectroscopy for chemical compound verification, high-resolution optical microscopy to identify diagnostic structural features like Schreger lines, as well as genetic profiling and carbon dating \cite{Burrgato1998, Cartier2020, Shepherd2024}. However, a major issue in the regular identification of original ivory is that these physical testing methods require direct access to the specimen, specialized laboratory hardware, and expert interpretation, rendering them impossible to deploy reactively at the point of digital advertisement.
In this context, a tool that would automatically classify ivory and bone in images from online adverts would be crucial to reduce the workload for the law enforcement officers, leading to an increased number of successful prosecutions.

Initial computational attempts to track the illegal wildlife trade on digital platforms have predominantly focused on textual data mining and automated keyword filtering. 
For instance, early frameworks utilized text-based data-mining algorithms to isolate suspicious metadata parameters on e-commerce platforms like eBay \cite{hernandez2015}, while other models applied unsupervised natural language processing, such as the biterm topic model, to detect concealed code words within social media text streams like Twitter \cite{Minin2019}. 
A significant limitation of these text-centric methods, however, is their vulnerability to evolving seller tactics; illicit traders frequently circumvent text filters by deliberately misspelling keywords, utilizing visual text-in-image descriptions, or omitting text descriptors entirely. While these studies highlighted the long-term value of integrating image-recognition models, practical applications utilizing direct pixel-level computer vision pipelines on photographic listings have remained limited (with prominent examples including Refs.~\cite{Xu2019, Kulkarni2023, Kulkarni2025}). 
The present work addresses this critical gap by evaluating an image-based segmentation approach capable of extracting visual evidence directly from the listing photograph itself. Specifically, in this work we investigate the potential of AI-based image analysis techniques applied to classification of ivory images using artificial intelligence (AI). A number of neural network architectures were trialled and the paper presents the best performing as a proof of concept. Specifically, we apply image segmentation methods in order to classify ivory and bone images as seen in online marketplaces. A neural network approach based on the U-Net architecture \cite{unet} is used to obtain a pixel level classification for ivory and bone objects. The code use in our approach is made available on GitHub\footnote{The code is available at \href{https://github.com/AndriusT/AIvory}{https://github.com/AndriusT/AIvory}.}. 

This paper is organized as follows: in the Dataset section we present the data used; the Method section describes how semantic image segmentation can be used to identify ivory, and we describe the particular model which we have implemented. In the Results section we demonstrate the accuracy of our approach on a test dataset which includes images of bone and ivory carvings. Finally, in the Discussion section we discuss the results, point out the limitations of our method and suggest future developments.

\section*{Dataset}
\label{Section:data}

\subsection*{Initial Dataset}
\label{Section:initial-dataset}
\noindent A total of 541 images were used to prepare the datasets, with 410 allocated to training and validation along with 131 allocated to testing. The training images include the following:

\begin{itemize}
    \item 181 are digital photographs of elephant ivory tusks and carved ivory. 85 of these were provided by the School of Law in the University of Portsmouth as a part of the University of Portsmouth Ivory Project collection. These particular images have a resolution of 4928$\times$3264 pixels. In addition, 96 of the images were newly photographed using a Canon PowerShot SX540 HS camera. These images have a resolution of 5184$\times$3456 pixels. All of the mentioned images are in JPG format.

    \item 106 are digital photographs of items resembling carved ivory. These objects are made from bone, horn or plastic and can be legally sold. Of these, 31 images were provided by the School of Law of the University of Portsmouth, 75 were new images photographed using a Canon PowerShot SX540 HS camera. The resolution of the mentioned images is 5184$\times$3456 pixels. The images are in the JPG format.

    \item A total of 123 images were obtained from Wikimedia Commons \cite{wikimedia} under a public domain license. These comprise 41 images of ivory and 82 images of bone objects. The images are provided in a range of file formats and resolutions.

\end{itemize}

The used images contain a wide selection of objects ranging from tusks to ivory and bone tools, jewellery, animal and human figurines and other museum exhibits. Some of the ivory objects show clearly visible Schreger lines while others do not. Objects are of different sizes and the photographs were taken under different lighting conditions with different color backgrounds (see Figs~\ref{Fig:sem_seg}, \ref{Fig:ivory} and \ref{Fig:bones}). This factor, in particular, makes it more difficult to train our algorithm; however, the diversity of the images makes our dataset more realistic and more akin to the images displayed in online marketplaces.

While this diversity supports our proof-of-concept approach, we acknowledge important differences between our dataset and real marketplace listings. Our images predominantly comprise of professionally photographed specimens, whereas sellers often use poor lighting, non-standard framing, and obfuscation tactics. The dataset size (541 images) reflects a practical constraint: we exhausted all publicly available ivory and bone images in Wikimedia Commons with appropriate licensing, supplemented by institutional and newly photographed specimens. The heterogeneous sources and class imbalance may introduce subtle biases, but the dataset is appropriate for transfer learning-based proof-of-concept work. Operational deployment will require substantially larger datasets (10,000+ images) from marketplace sources, including images with poor lighting conditions and photographs used for deliberate misrepresentation.

\subsection*{Image Augmentation}
\label{Section:augmentation}

Note that the numbers listed above (i.e., 410 images used for training) refer only to the number of original images in the dataset and do not include those generated through image augmentation. Image augmentation refers to the technique of increasing the dataset size by applying various transformations to the images, such as cropping, rotating, reflecting, and resizing.

In our case, we applied a number of different augmentation strategies. In particular, in order to explore the effectiviness of different augmentation methods, we generated four distinct versions of the original training dataset, which were then used to train four distinct machine learning models. These models were as follows:

\begin{itemize}
    \item Model I: No augmentation. Each image is resized to $400 \times 400$ pixels.

    \item Model II: Images are resized to $400 \times 400$ pixels and subjected to a standard 
    augmentation pipeline provided by the {\tt fastai} library \cite{fast}: random horizontal 
    flips; rotations drawn uniformly from $[-10^{\circ}, +10^{\circ}]$; zoom factors drawn uniformly from 
    $[1.0, 1.1]$; random perspective warping with a maximum warp coefficient of $0.2$; and 
    random brightness/contrast adjustments with a maximum lighting change of $0.2$. 
    The affine and lighting transforms are each applied independently with probability $0.75$.

    \item Model III: same as Model II, but {\tt Resize} is replaced by {\tt RandomResizedCrop}: 
    a rectangular region is sampled uniformly such that it covers at least $30\%$ of the 
    original image area and with a random aspect ratio, and is then resized to $400 \times 400$ 
    pixels. This forces the network to be invariant to both scale and spatial location of the 
    subject within the frame.

    \item Model IV: same as Model III, with two additional colour augmentation strategies applied on top of the standard pipeline: random hue shifts drawn uniformly 
    from $[-0.25, +0.25]$ (on a scale where $1.0$ corresponds to a full $360^{\circ}$ rotation of 
    the colour wheel, i.e.\ up to $\pm 90^{\circ}$), and random saturation adjustments with a 
    maximum factor of $0.5$. These colour augmentations are intended to make the classifier 
    robust to variations in photographic lighting conditions and camera colour balance.
\end{itemize}

With the listed augmentation methods applied to a random subset of images during each epoch of the training procedure, the effective number of images seen by the algorithm is then in the range of 2500--3000 images given the settings we used for training (see the Model subsection). Note that the range of obtained images corresponds to the stochastic nature of the listed augmentation procedures resulting in a different number of total images, depending on which particular data samples were chosen to be in the training dataset.

\section*{Method}
\label{Section:method}

Ivory-bone object classification requires two separate steps: to identify the location of the item in the image, and then to classify the object as ivory or not ivory. This approach needs to be generalised in case multiple objects are present in the image.

\begin{figure}[!h]
  \centering
  \includegraphics[width=0.35\textwidth, height=7cm]{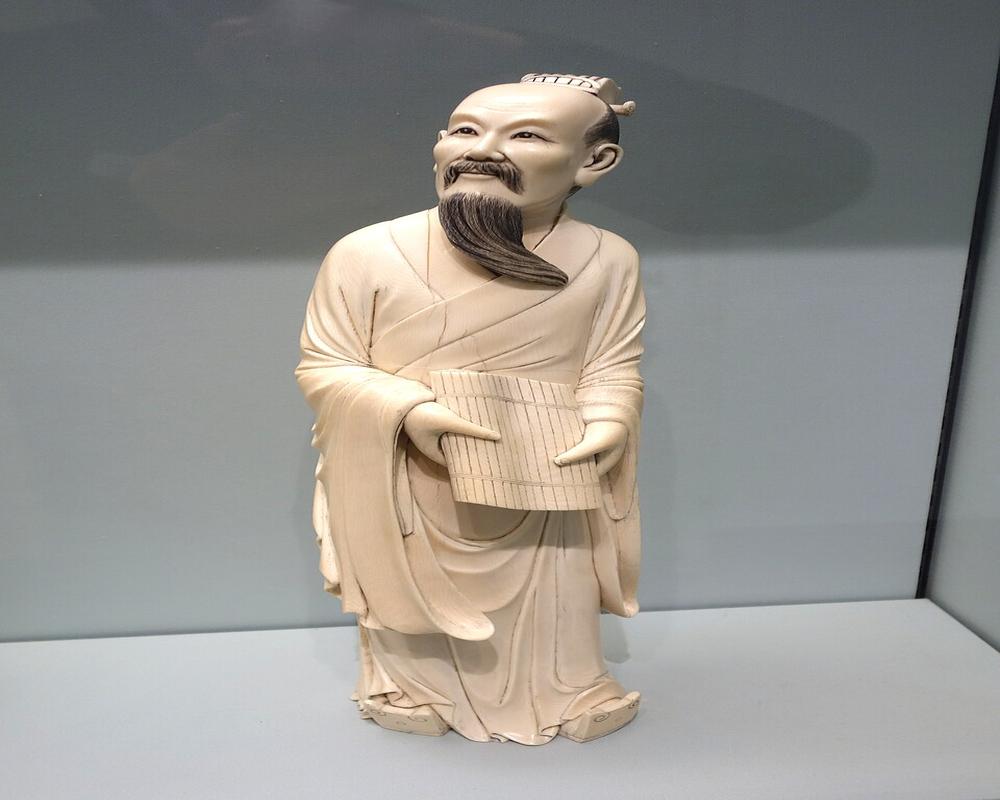}\includegraphics[width=0.35\textwidth, height=7cm]{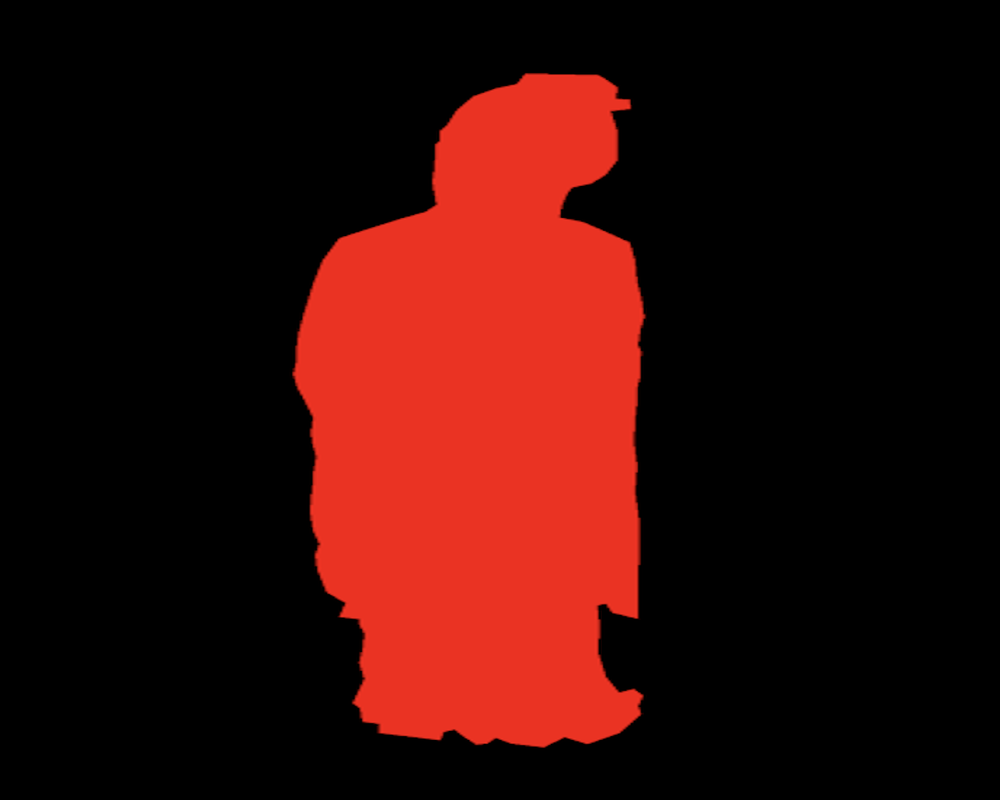}
  \includegraphics[width=0.35\textwidth, height=7cm]{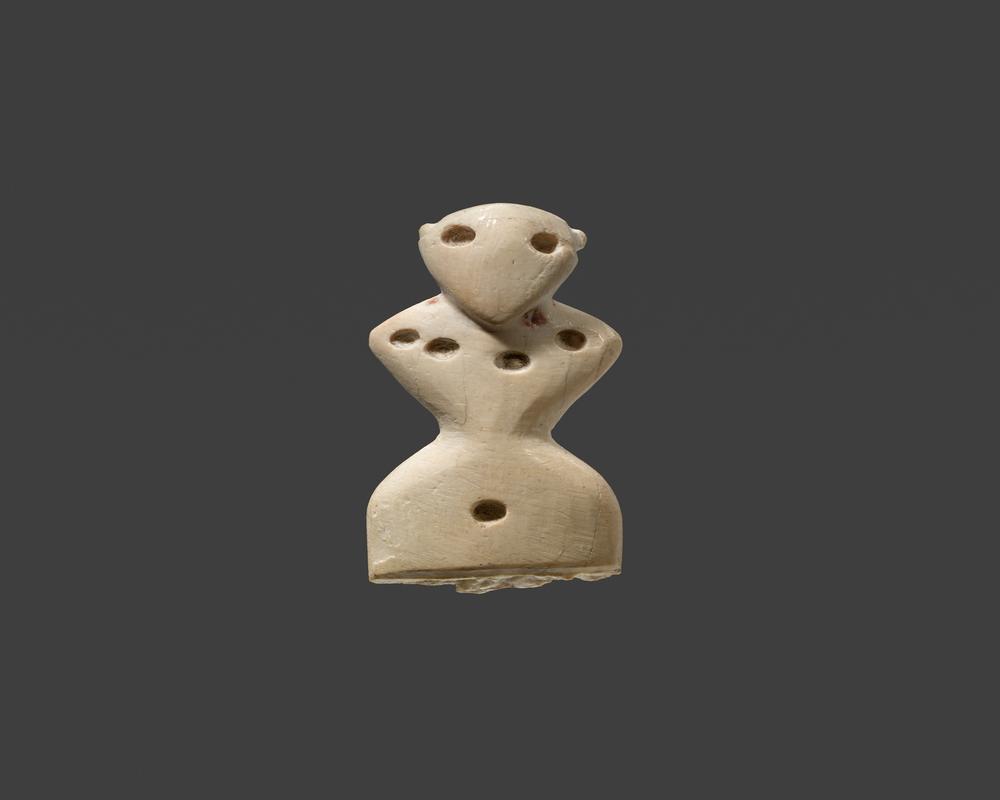}\includegraphics[width=0.35\textwidth, height=7cm]{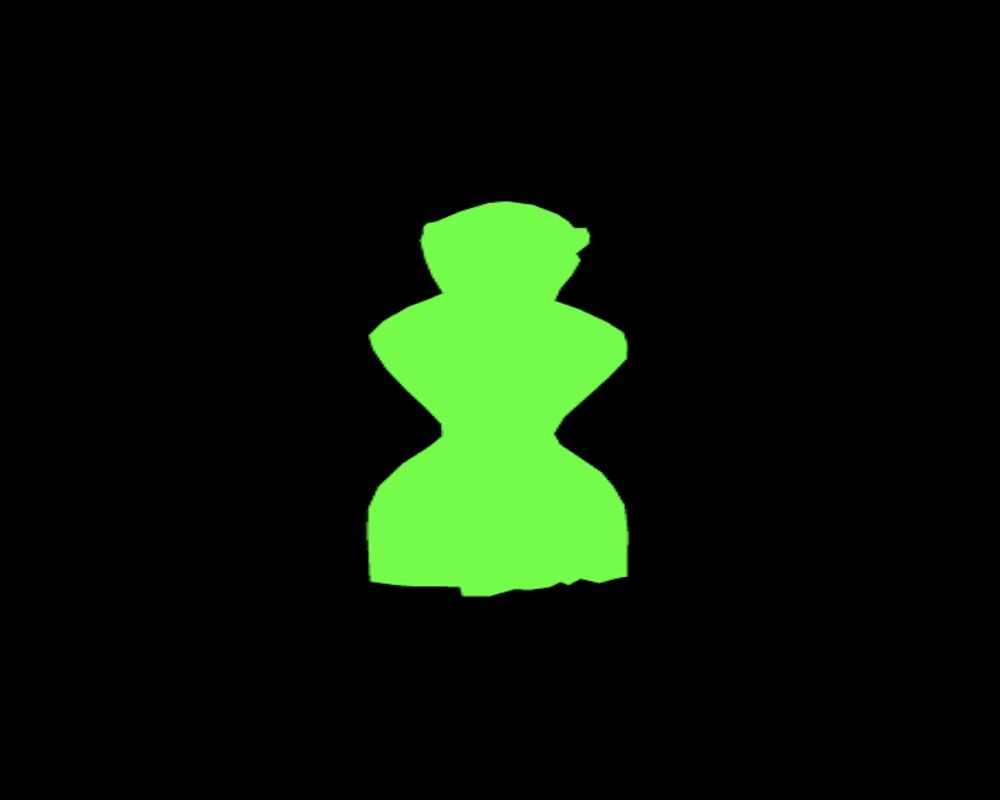}
  \caption{\textbf{Example of semantic segmentation.} The original images are shown on the left, while the images on the right show the classification targets. Red pixels correspond to the ivory class, green pixels to non-ivory, and black pixels to the background (images from Wikimedia Commons under the public domain license).}
  \label{Fig:sem_seg}
\end{figure}

In computer vision, this problem can be addressed with a semantic image segmentation approach (e.g., see \cite{Minaee2022} for an overview). Semantic segmentation is the process of assigning for each pixel in an image the class to which it belongs; hence it is equivalent to a classification problem per pixel. Fig~\ref{Fig:sem_seg} shows an example of semantic segmentation for 2 pictures from our dataset. From the lower images we notice that this approach does not distinguish instances of the same class, i.e. two objects which belong to the same category will be labelled in the same way.

The benefit of semantic segmentation is that it does not explicitly require identifying the location of the object in the image. On the downside, training an image segmentation model is time-consuming because it requires manually labeling the available images. For each image in our dataset, we created a ``mask" similar to the one shown in Fig~\ref{Fig:sem_seg} using the graphical image annotation tool {\tt labelme}\footnote{\url{https://github.com/wkentaro/labelme}.} \cite{wada_labelme}. Finally, each pixel of each image was then classified as ``ivory", ``non-ivory" or ``background" using the U-net based approach described in the Model subsection.

Our choice of transfer learning and semantic segmentation was informed by the dataset limitations outlined above. Transfer learning leverages pre-trained features from ImageNet \cite{deng2009imagenet}, mitigating the impact of our relatively small training set and heterogeneous image sources by reducing the effective domain shift between museum-quality images and marketplace listings. Semantic segmentation at the pixel level allows the model to identify discriminative features (e.g., Schreger lines) regardless of image quality or background variation, providing robustness across source heterogeneity. However, the class imbalance in our training data (roughly 63\% ivory, 37\% bone) may bias predictions toward the majority class. To account for this, we employ a threshold-based classification scheme (see section Results) that allows the model to express uncertainty, providing a more nuanced assessment than hard binary classification.

\subsection*{Model}
\label{model-section}

\begin{figure}[!h]
  \centering
  \includegraphics[width=1.\textwidth]{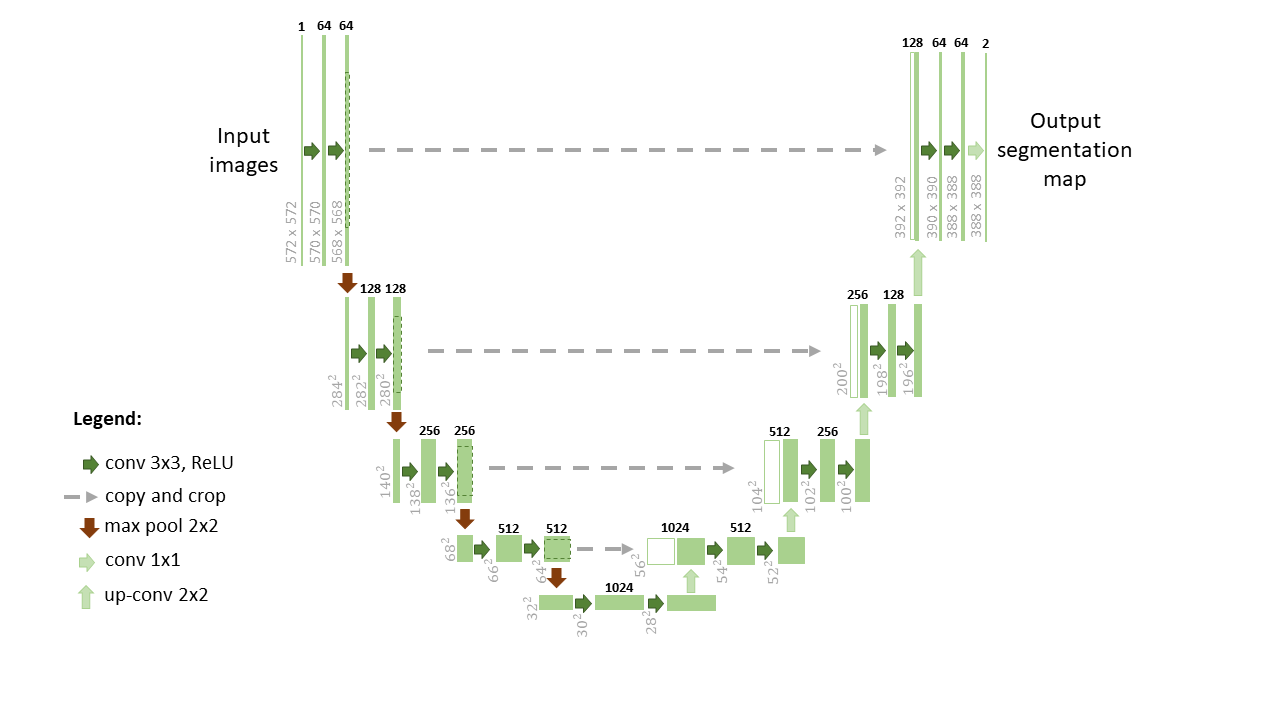}
\caption{\textbf{The U-Net architecture used for image segmentation.} The U-net architecture consists of a contracting path (left side) made of down-sampling units, followed by an expansive path (right side) consisting of up-sampling units. Each down-sampling unit consists of convolution layers that extract features from the input images, and a max-pooling layer to reduce the size of the feature map. The extracted features are passed down to the next down-sampling unit and the corresponding up-sampling unit.}
\label{Fig:unet}
\end{figure}

We create a framework for semantic segmentation using the U-net model \cite{unet} with Resnet34 architecture \cite{he2015deep} as backbone. Fig~\ref{Fig:unet} shows our architecture, which is similar to the first U-net originally developed for biomedical image segmentation \cite{unet}. This particular model was chosen due to its proven track record with small datasets (e.g. $\mathcal{O}(1,000)$ images). To improve the performance and to reduce the training time the first half of the U-net (encoder) was initialised using weights pre-trained over the ImageNet\footnote{\url{https://www.image-net.org/}.} dataset. This approach is called ``Transfer Learning" \cite{zhuang2020comprehensive} and it generally helps with saving on computational, financial and environmental costs related to the training procedure.

We develop our model using {\tt fastai} and its DynamicUnet class \cite{fast}. {\tt fastai} is a high-level API on top of PyTorch \cite{NEURIPS2019_9015}. The DynamicUnet class automatically creates a model adapting the intermediate feature sizes in accordance with the image size.

We trained the U-Net using only the ivory and bone images, after splitting the dataset into 80\% for training and 20\% for validation. We trained the model for 20 epochs using Google Colaboratory\footnote{\url{https://colab.research.google.com/}.}, which is a cloud-based Jupyter notebook environment that lets users execute code directly through a web browser with access to free GPU and TPU compute resources. The number of training epochs was determined by trial and error -- we found 20 epochs to be sufficient for the model to converge in all considered cases. Using various image resizing and augmentation strategies we reach accuracies ranging between 90\% and 95\% on the training dataset, where the accuracy is defined as the average fraction (arithmetic mean) of pixels, which do not belong to the background and that are well classified.
In particular, we classify carvings in the images based on the model average prediction $\hat{p}$, estimated as:
\begin{equation}\label{onlyeq}
    \hat{p} = \sum_{pixels}\frac{p_{ivory}-p_{bone}}{p_{ivory}+p_{bone}}~,
\end{equation}
where $p_{ivory / bone}$ is the probability of a pixel to belong to the ivory/bone class and the summation excludes pixels identified as background. 
By definition, $\hat{p}$ can assume values between $-1$ and $+1$, with negative values indicating that the object is more likely to be made of bone, while positive values suggest the presence of ivory. For each image, we report the fraction of pixels classified as ivory, the average prediction $\hat{p}$ and the predicted class. We classify each item as ivory if $\hat{p}>0.3$ or bone if $\hat{p}<-0.3$. Images with $|\hat{p}| < 0.3$ are classified as ``uncertain", indicating that the model cannot clearly determine the material of the item.

\section*{Results}
\label{Section:results}

We tested our model using a set of 131 unseen images, taken from the publicly available archive at Wikimedia Commons \cite{wikimedia}. This set consists of 73 images of carved ivory and 58 images of bone objects, photographed using different setups and under different lighting conditions. As before, the number of images was increased to around 800 via image augmentation. The photography information was generally not provided by the owner (e.g., metadata; however, note that metadata is not used in any way in our approach). The identification labels of each image (ivory or bone) were double-checked by an expert in elephant ivory identification, and only the confirmed images were used in the final dataset.

\begin{figure}[!h]
  \centering
  \includegraphics[width=0.6\textwidth]{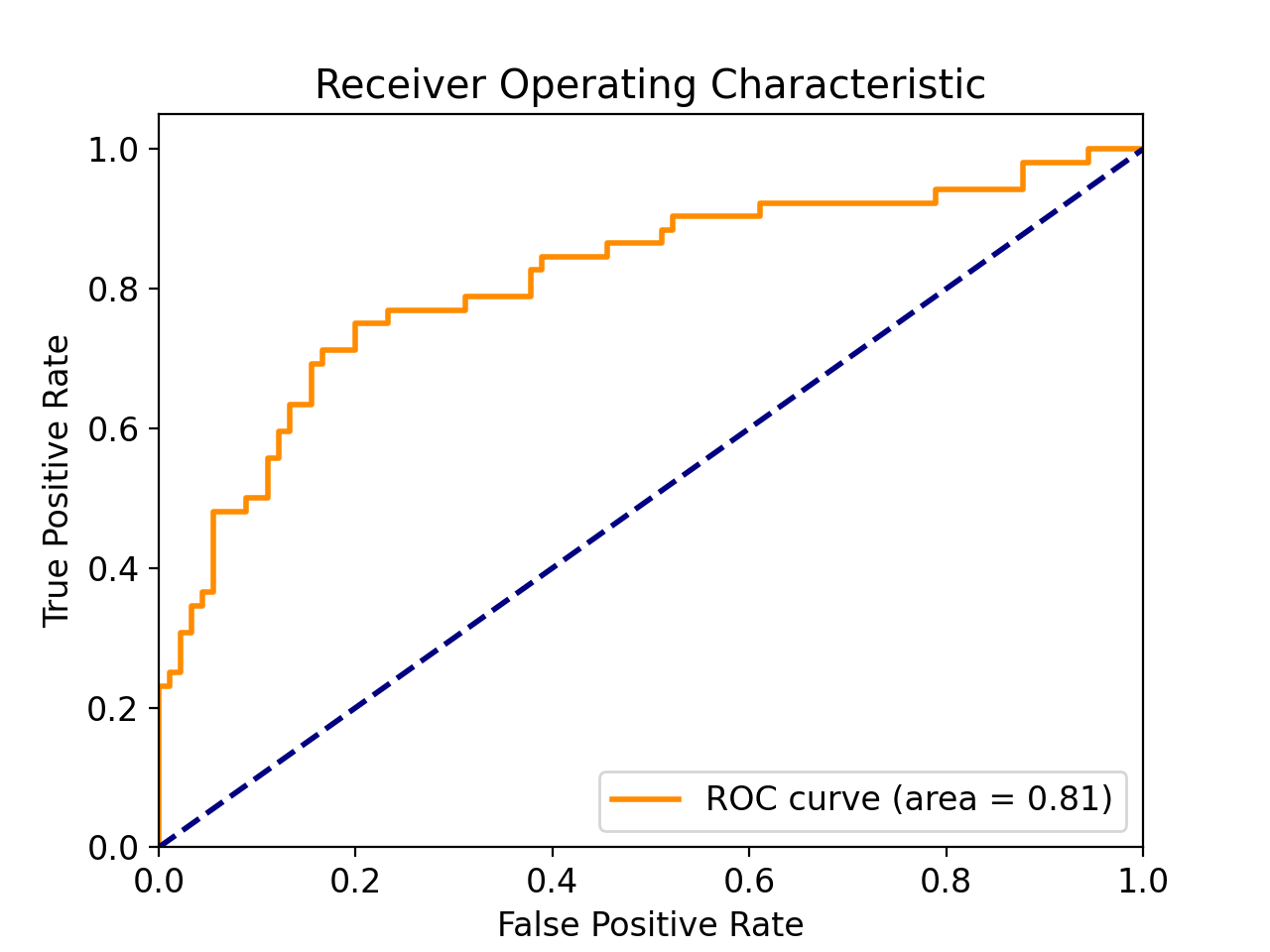}
\caption{\textbf{Receiver operating characteristic (ROC) curve for the ivory classifier.} The area under the ROC curve is 0.81. The model performs significantly better than a random classifier (shown by the dashed line).}
\label{Fig:ROC}
\end{figure}

Fig~\ref{Fig:ROC} shows the receiving operating characteristic (ROC) curve for the ivory classifier. The ROC curve evaluates the performance of the classifier for each classification threshold, i.e. choosing different $\hat{p}$ values to distinguish ivory carvings from bone. In particular, the area under the ROC curve (AUC) presents an aggregate measure of the classifier performance and can be interpreted as the probability that the model ranks ivory more highly than bone. 

\begin{figure}[!h]
\centering
\includegraphics[height=0.4\paperheight]{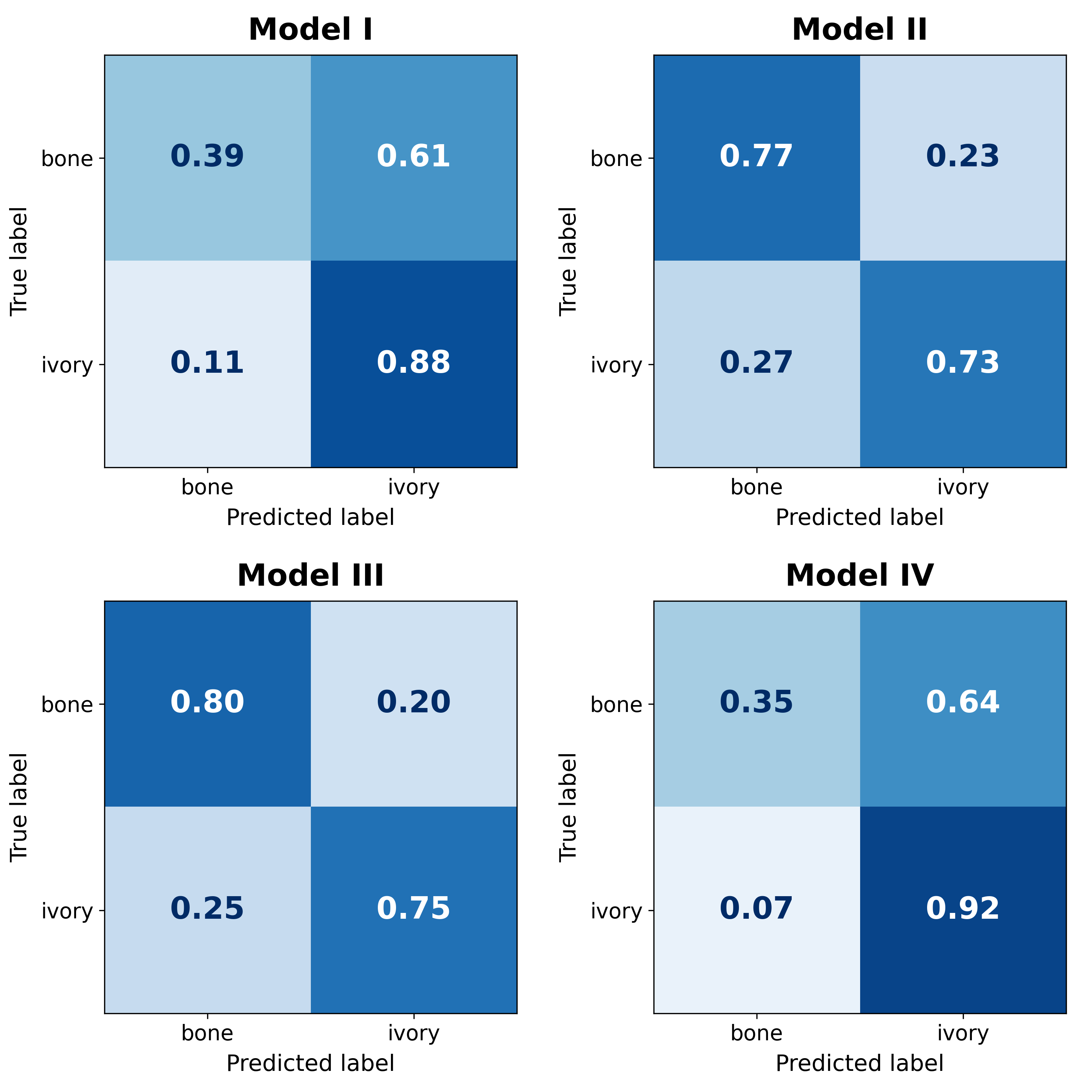}
\caption{The confusion matrices for the different models (different image augmentation strategies) evaluated on the full test dataset.}
\label{Fig:CM-full-test-data}
\end{figure}

\begin{figure}[!h]
\centering
\includegraphics[height=0.4\paperheight]{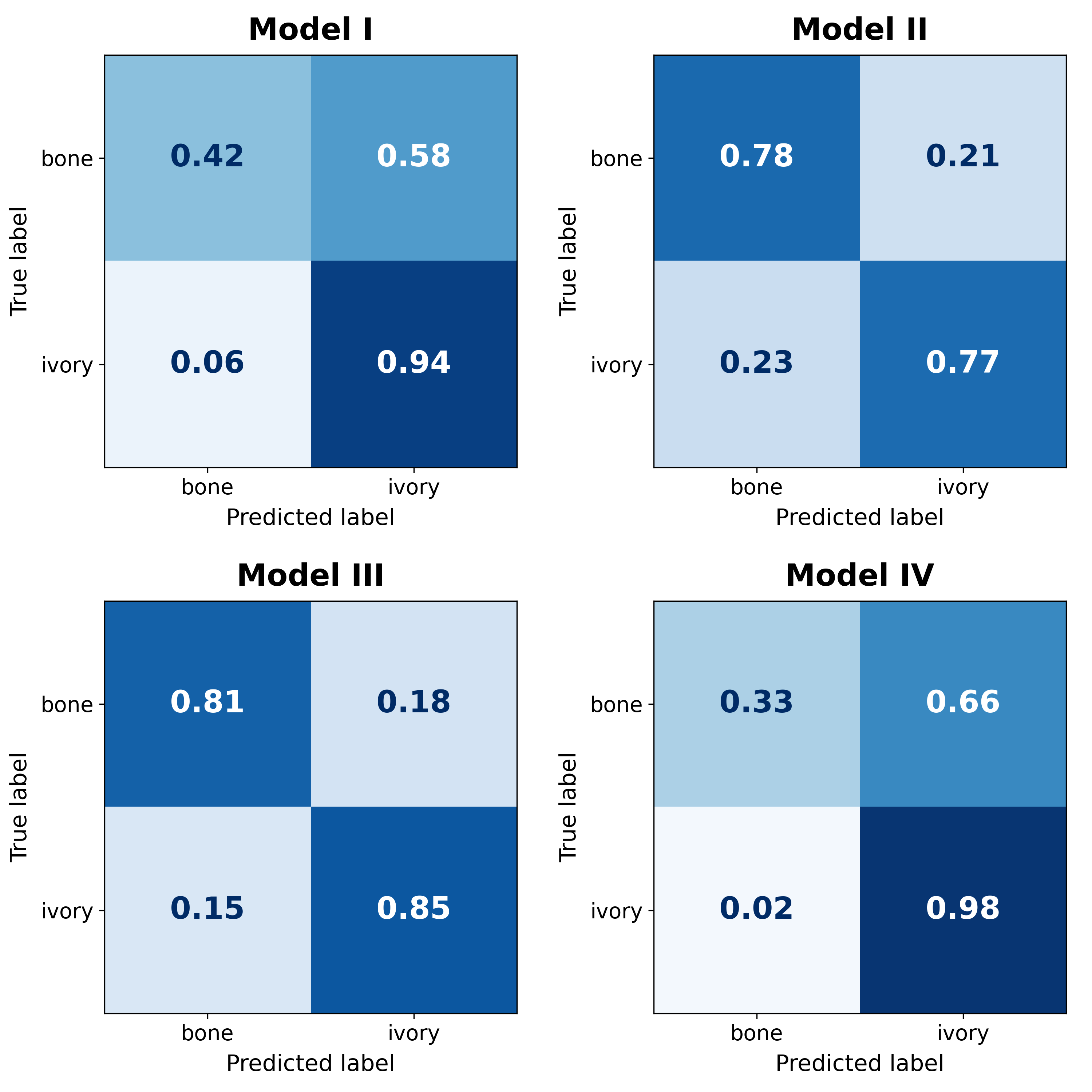}
\caption{The confusion matrices for the different models evaluated for confident predictions ($|\hat{p}| > 0.3$) only.}
\label{Fig:CM-confident-predictions}
\end{figure}

In addition, we report other statistics of interest, including precision, recall, F1-score, area under the curve (AUC) and the Youden's J statistic. The mentioned statistics are provided for the full test dataset in Table \ref{table:full-test-statistics}. In Table \ref{table:confident-test-statistics}, we provide the listed statistics for the test predictions that are categorized as \textit{confident}. This is defined as predictions for which the averaged ivory probability excess is higher than a chosen threshold value, i.e., $|\hat{p}| > 0.3$. We find that the best-performing model's AUC value is 0.81 for the full dataset and 0.86 for the confident predictions dataset. We also provide the confusion matrices for the full test dataset and the confident predictions in Fig. \ref{Fig:CM-full-test-data} and \ref{Fig:CM-confident-predictions} correspondingly.

\begin{figure}[!h]
  \centering 
  \includegraphics[height=0.78\paperheight]{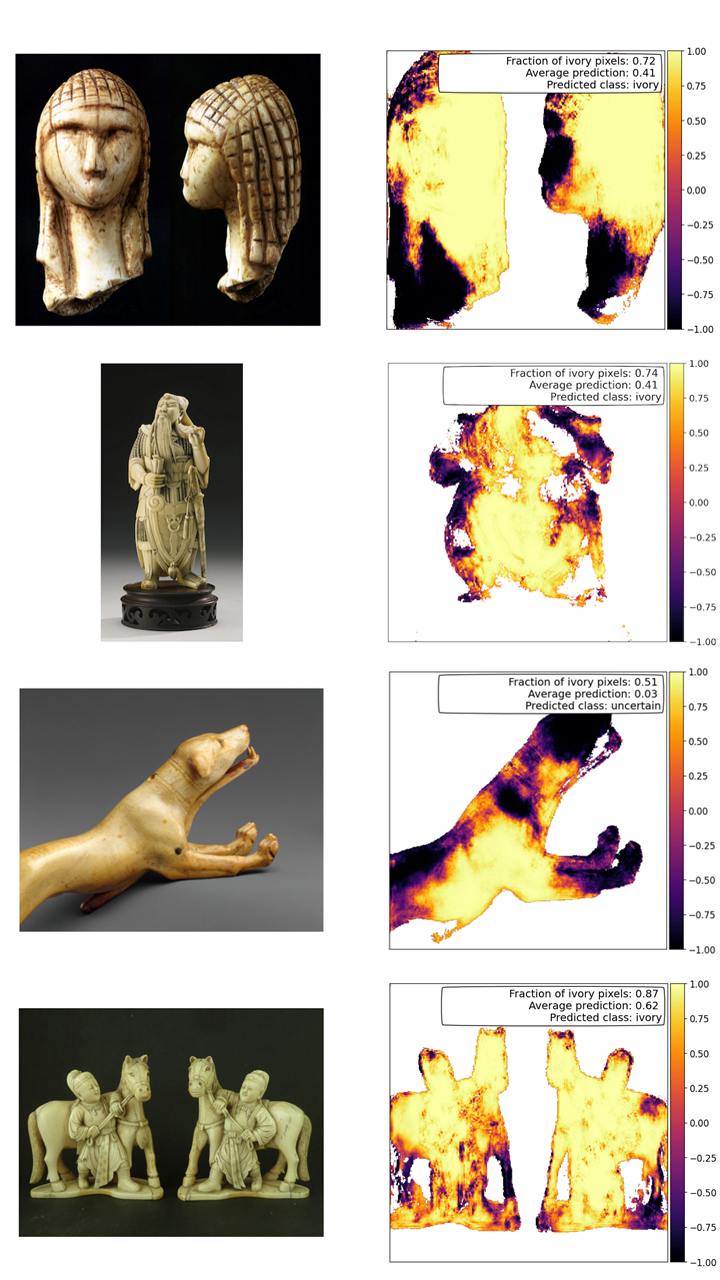}
\caption{A selection of carved ivory images along with their pixel-level classification. The colors indicate the pixel classification (darker for bone, lighter for ivory and white for background). The figure legends also show the predicted class and the fraction of pixels classified as ivory.}
\label{Fig:ivory}
\end{figure}

\begin{figure}[!h]
  \centering
  \includegraphics[height=0.78\paperheight]{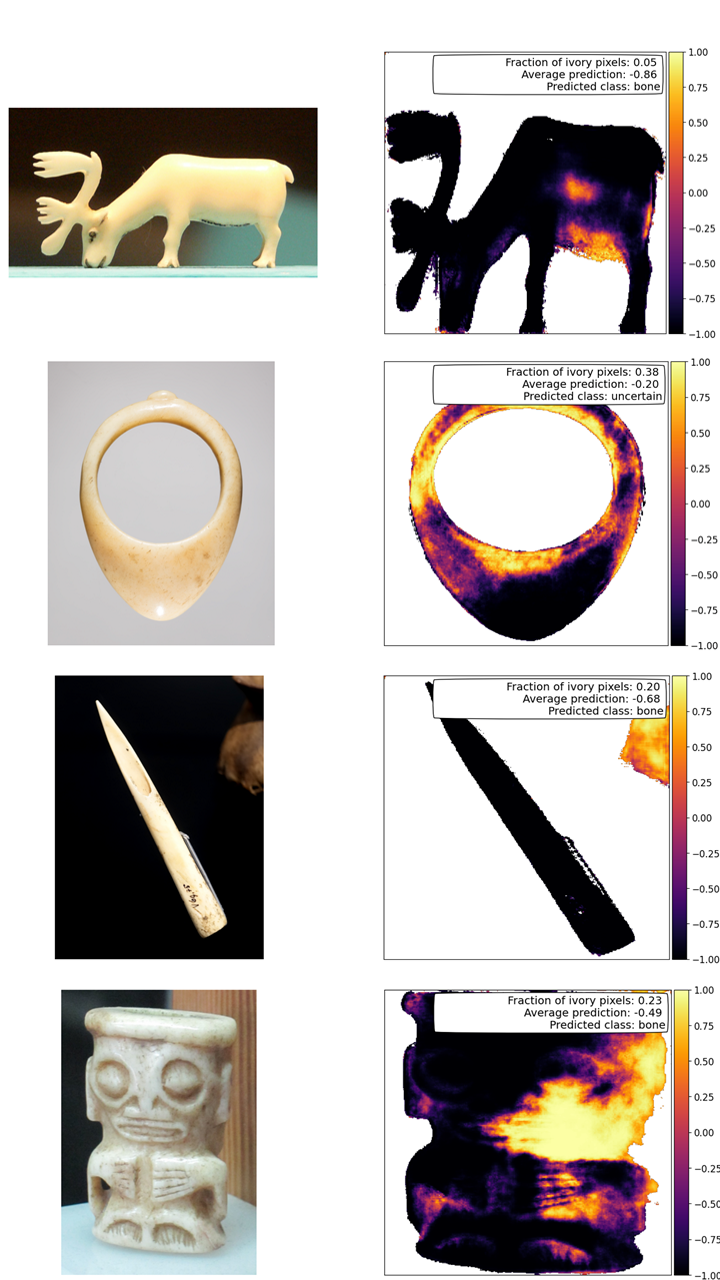}
\caption{A selection of bone images from our test dataset. The colorbar and legend conventions are the same as in Fig~\ref{Fig:ivory}.}
\label{Fig:bones}
\end{figure}

Figs~\ref{Fig:ivory} and \ref{Fig:bones} show the predictions for a sample of carved ivory and bone images from the test set. The colorbar values are related to the probability assigned to each pixel of belonging to each class (i.e. ivory, bone, or background). Specifically, a lighter color indicates that a given pixel is more likely to be identified as ivory, while darker colors refer to the bone class. Similarly, white pixels are assigned to the background class. Note that the model accurately identifies the targeted objects in all images, even when multiple items are present. The background is correctly segmented, even when it is non-uniform, as illustrated in the first example of Fig~\ref{Fig:ivory}.

\begin{table}[h]
\centering
\caption{Model performance metrics for ivory classification for different image augmentation strategies. The statistics here were calculated including all the images in the test dataset (those deemed certain and uncertain by the algorithm). See the Dataset section for a full description of the different models. The best performing model is marked in bold.}
\label{tab:model_metrics}
\begin{tabular}{lccccc}
\toprule
\textbf{Model} & \textbf{Precision} & \textbf{Recall} & \textbf{F1-Score} & \textbf{AUC} & \textbf{Youden's J} \\
\midrule
Model I & 0.46 & 0.88 & 0.60 & 0.75 & 0.43 \\
Model II & 0.64 & 0.73 & 0.68 & 0.77 & 0.52 \\
\textbf{Model III} & \textbf{0.68} & \textbf{0.75} & \textbf{0.72} & \textbf{0.81} & \textbf{0.55} \\
Model IV & 0.45 & 0.92 & 0.61 & 0.80 & 0.50 \\
\bottomrule
\end{tabular}
\label{table:full-test-statistics}
\end{table}

\begin{table}[h]
\centering
\caption{Model performance metrics for ivory classification for different image augmentation strategies. In this case we only consider test predictions that were marked as \textit{confident} (i.e., for which the averaged ivory probability excess score was higher or equal to 0.3). The best-scoring model is marked in bold.}
\label{tab:model_metrics}
\begin{tabular}{lccccc}
\toprule
\textbf{Model} & \textbf{Precision} & \textbf{Recall} & \textbf{F1-Score} & \textbf{AUC} & \textbf{Youden's J} \\
\midrule
Model I & 0.68  & 0.94  & 0.79  & 0.68  & 0.35   \\
Model II & 0.67   &  0.77  & 0.71   & 0.79  & 0.55   \\
\textbf{Model III} & \textbf{0.82}  & \textbf{0.85}   & \textbf{0.84}  & \textbf{0.86} & \textbf{0.67}  \\
Model IV & 0.57  & 0.98  & 0.72  & 0.78  & 0.49 \\
\bottomrule
\end{tabular}
\label{table:confident-test-statistics}
\end{table}

For each picture, we report the fraction of pixels classified as ivory, the average prediction $\hat{p}$ and the predicted class. We classify each item as ivory if $\hat{p}>0.3$ or bone if $\hat{p}<-0.3$. As before, images with $|\hat{p}| < 0.3$ are classified as ``uncertain", indicating that the model cannot clearly determine the material of the item.
Setting these thresholds, the model predicts the correct class for 59\% of the images, the wrong class for 14\%, and is uncertain for the remaining 27\% of the cases. For test images on which the algorithm is confident (for the best scoring model), this corresponds to an overall classification accuracy of 83\%. The threshold values for $\hat{p}$ were determined experimentally; it is therefore possible that improved results could be achieved through hyperparameter tuning, which we leave for future work.

In general, the algorithm gives mostly accurate predictions for the items that clearly show evidence of Schreger lines (e.g. the bottom image in Fig~\ref{Fig:ivory}). However, as shown in the third example of Fig~\ref{Fig:ivory}, the model tends to miscategorize objects which have bright spots, even if they present Schreger lines.
Note that even if the color of the objects does not seem to affect the classification results significantly, we notice a strong correlation between the prediction class and the lighting conditions in the image. The model has a tendency to classify darker objects as ivory, while well-lit pictures are often predicted to contain bone items.

The accuracy figures listed above are best interpreted in the context of human performance. For example, a classification accuracy of 81\% represents strong performance when compared to an ivory expert working without access to the objects themselves, relying solely on limited-resolution images and no additional information. A valuable direction for future work would be to have an independent ivory expert classify the test images in our dataset, providing an important benchmark against which to compare our results.

\section*{Discussion}
\label{Section:dev}

In this work, we have presented one of the first attempts to use computer vision techniques to identify ivory items solely from pictures (other similar approaches include \cite{Castro_2015, Xu2019}). Our goal is to develop a tool that could help law enforcement officers to identify illegal trading of ivory in online marketplaces. The task is extremely challenging due to the great variety of ivory carvings and the presence of materials that resemble ivory. Moreover, being a biological material, ivory tends to change in color with age and exposure to light.

We addressed the problem using a semantic image segmentation approach and we focused on distinguishing ivory carvings from bone items. Our model shows a good predictive ability when evidence of the type of material (e.g. Schreger lines) is visible.
However, we identify a clear dependence of the classification accuracy on the lighting of the item in the image. 

The statistical analysis performed on the test dataset (Tables \ref{table:full-test-statistics} and \ref{table:confident-test-statistics}) also provides valuable insight into promising image augmentation strategies. First, the primary challenge to our approach is the small size of the training dataset. This is illustrated by the underperformance of Model I in Tables \ref{table:full-test-statistics} and \ref{table:confident-test-statistics}. Second, different augmentation strategies significantly improve almost all of the considered statistics; however, the type of augmentation strategy employed matters. In particular, the best-scoring model, Model III, implements a wide range of augmentation transformations, including random rotations, zooms, warping transformations, brightness and contrast adjustments, and random crops. However, when random hue and saturation transformations are added to this set (Model IV), the algorithm predicts ivory in most cases. This provides another valuable indication that lighting conditions and colour scheme are among the key features that the algorithm uses to distinguish ivory from bone. Overall, these results reaffirm the importance of training dataset size and image diversity. The issues identified here could be further mitigated by obtaining a larger dataset, as well as by employing generative AI methods to perform additional image augmentation.

It is also important to discuss our results in a wider context of different detection and classification of ivory techniques. 
Traditional approaches to ivory identification rely on destructive or access-dependent methods. Microscopy, for instance, identifies ivory through diagnostic morphological features—most notably Schreger lines, but requires direct physical access to specimens and expert interpretation. Spectroscopy-based approaches (primarily Fourier-transform infrared spectroscopy and Raman spectroscopy) provide definitive chemical identification and are non-destructive, yet similarly demand in-person specimen analysis and specialized laboratory infrastructure. By contrast, our computer vision approach operates solely on photograph evidence, enabling rapid screening of online marketplace listings where physical access is impractical. However, this strength comes with limitations: our study focuses on distinguishing processed ivory carvings from bone objects in relatively high-quality images. Real-world marketplace listings often feature poor lighting, obstructed views, and deliberate misrepresentation that may degrade performance compared to laboratory-quality specimens. Additionally, our current model cannot differentiate between freshly carved ivory and aged historical pieces, or identify ivory from specific elephant populations—capabilities achievable with spectroscopic methods. These limitations underscore that image-based AI classification is best positioned as a complementary first-pass screening tool rather than a definitive replacement for traditional forensic methods.

Despite the challenges, our model demonstrates solid predictive ability (for such a small dataset size), with potential for further improvement through enhanced tuning.
We are now in a position to try different machine learning models and approaches, which could improve performance, but a greater benefit is likely to come from using a larger and more heterogeneous set of images to train the model. With a larger dataset ($\mathcal{O}(10,000)$ images) it could also be possible to attempt a multi-class segmentation, enabling the model to distinguish other materials which resemble ivory, such as horns or ivorine, and also to distinguish between various types of ivory.

Once the model is optimised, the performances could be further improved using complementary approaches. For example, Ref.~\cite{Castro_2015} shows promising results using data-mining algorithms to classify the textual information associated with the items offered for sale. Likewise, the recently developed OpenAI's CLIP framework allows training machine learning classifiers on text and image pair data simultaneously \cite{CLIP}.
Incorporating textual information would increase the robustness of the classification over variations in the way illegal articles are advertised online. Such a multimodal approach would have important practical implications both for law enforcement working in the area of wildlife crime and for the multi-national companies hosting online auction platforms, such as eBay. Currently, the eBay algorithm scans advertisements for the use of ``key words" such as ``ivory" but it does not have the ability to identify ivory being offered for sale on its platform by the photographs posted by the seller. As previous research has shown, those knowingly offering ivory for sale in contradiction of the laws of the UK and of eBay's own selling regulations, use pseudonyms to evade text analysis. The model discussed in this paper, based on photographic image interrogation, could make these kinds of deliberate attempts to circumvent laws and regulations much more difficult. The potential impact of this on conservation is considerable.
The use of AI in conservation is growing. Predictive modelling is seen being used in species distribution research and AI-powered technologies, such as sensors are being effectively used in wildlife conservation, monitoring and in anti-poaching efforts \cite{Nandutu2023}. Technological advancements have been shown to be a useful addition to the intelligence gathering tools used in `smart' law enforcement \cite{Rademacher2019} and it is both rational and pragmatic to foresee a use for the type of model examined in this paper in wildlife crime law enforcement. For officers, such a model will have profound time-saving implications when investigating ivory being sold through online marketplaces and social media site. However, in addition to the clear law enforcement applications, for online marketplaces and social media sites themselves, such a tool empowers them to control what is advertised for sale on their sites.

\clearpage
\section*{Acknowledgments}

We thank Domagoj Korais for the useful discussion and suggestions.
The authors are grateful for the computational resources provided by the University of Portsmouth and Google.

\section*{Conflict of interest}

The authors declare no conflict of interest.

\section*{Data availability statement}

The key scripts used to obtain the results presented in this paper are available in the GitHub repository \href{https://github.com/AndriusT/AIvory}{https://github.com/AndriusT/AIvory}. The dataset used in this work can be accessed on Zenodo at \href{https://doi.org/10.5281/zenodo.21922172}{https://doi.org/10.5281/zenodo.21922172}.

\nolinenumbers

\bibliography{references}

\end{document}